\documentclass[letterpaper, 10 pt, journal, twoside]{IEEEtran} 

\IEEEoverridecommandlockouts                              

\usepackage[utf8]{inputenc} 
\usepackage[T1]{fontenc}    
\usepackage{url}            
\usepackage{booktabs}       
\usepackage{amsfonts}       
\usepackage{nicefrac}       
\usepackage{xcolor}  

\usepackage{multirow} 
\usepackage{multicol}
\usepackage{graphicx}
\usepackage{enumitem}
\usepackage[font=footnotesize]{caption}
\usepackage[font={footnotesize}]{subcaption}
\usepackage{pifont}
\usepackage{cuted}

\definecolor{darkgreen}{RGB}{0, 150, 0}
\definecolor{darkred}{RGB}{200, 0, 0}

\newcommand{\yes}{{\color{darkgreen} \ding{51}}}
\newcommand{\no}{{\color{darkred} \ding{55}}}
\newcommand{\secref}[1]{Sec.~\ref{#1}}
\newcommand{\figref}[1]{Fig.~\ref{#1}}
\newcommand{\tabref}[1]{Tab.~\ref{#1}}

\newcommand\myworries[1]{\textcolor{black}{#1}}
\definecolor{cvprblue}{rgb}{0.21,0.49,0.74}
\usepackage[breaklinks,colorlinks]{hyperref}

\title{\LARGE \bf
AmodalSynthDrive: A Synthetic Amodal Perception Dataset for\\Autonomous Driving
}

\author{Ahmed Rida Sekkat$^{*,1}$, 
        Rohit Mohan$^{*,2}$,
        Oliver Sawade$^{1}$,
        Elmar Matthes$^{1}$,
        and~Abhinav Valada$^2$
\thanks{$^*$These authors contributed equally.}%
\thanks{$^1$IAV GmbH, Germany}%
\thanks{$^2$Department of Computer Science, University of Freiburg, Germany}
\thanks{The Supplementary material is available on arXiv at \url{https://arxiv.org/abs/2309.06547}.}
}

\begin{document}

\maketitle
\thispagestyle{empty}
\pagestyle{empty}

\begin{abstract}
Unlike humans, who can effortlessly estimate the entirety of objects even when partially occluded, modern computer vision algorithms still find this aspect extremely challenging. Leveraging this amodal perception for autonomous driving remains largely untapped due to the lack of suitable datasets. The curation of these datasets is primarily hindered by significant annotation costs and mitigating annotator subjectivity in accurately labeling occluded regions. To address these limitations, we introduce AmodalSynthDrive, a synthetic multi-task multi-modal amodal perception dataset. The dataset provides multi-view camera images, 3D bounding boxes, LiDAR data, and odometry for 150 driving sequences with over 1M object annotations in diverse traffic, weather, and lighting conditions. AmodalSynthDrive supports multiple amodal scene understanding tasks including the introduced amodal depth estimation for enhanced spatial understanding. We evaluate several baselines for each of these tasks to illustrate the challenges and set up public benchmarking servers. The dataset is available at \url{http://amodalsynthdrive.cs.uni-freiburg.de}.
\end{abstract}

\section{Introduction}

The process of amodal perception in human cognition allows us to perceive and understand complete objects from partial or obscured visual stimuli, which is fundamental for our interaction with the surrounding world~\cite{goldstein2021sensation,smith2013cognitive}. 
This perceptual ability, inherent in our cognitive processes, bears significant promise for advancing the capabilities and safety measures of autonomous systems, especially in regard to autonomous driving. The incorporation of amodal perception tasks within these systems can boost their understanding of complex environments through improved occlusion reasoning and depth ordering for a more holistic environmental perception~\cite{mohan2023panoptic, mohan2023neural}.
This enhanced perception can assist in precise object tracking~\cite{buchner20223d}
, SLAM
~\cite{vodisch2022continual}
, and informed decision-making~\cite{schmalstieg2022learning}. In recent years, there has been an increasing interest in amodal perception~\cite{li2016amodal, tran2022aisformer, breitenstein2022amodal, mohan2022amodal, mohan2022perceiving}. Despite the inherent ill-posed nature of predicting occluded regions, studies have shown that humans excel and benefit from this for their cognitive understanding~\cite{zhu2017semantic}. However, we are yet to model this amodal perception ability in autonomous systems due to the lack of high-quality annotated data, which presents a significant challenge for advancing research in this area.
Real-world sensory data capturing the entire structure of occluded object regions in diving scenes is difficult to obtain, as vehicle sensors cannot observe these occluded regions. This limitation renders the task of pixel-level labeling by human annotators both demanding and subjective, potentially introducing bias. Consequently, the data curation process becomes prohibitively expensive, exacerbated by the extensive efforts needed to reduce annotator subjectivity in accurately labeling occluded regions.\looseness=-1

\begin{figure}[!t]
    \centering
        \includegraphics[width=\linewidth]{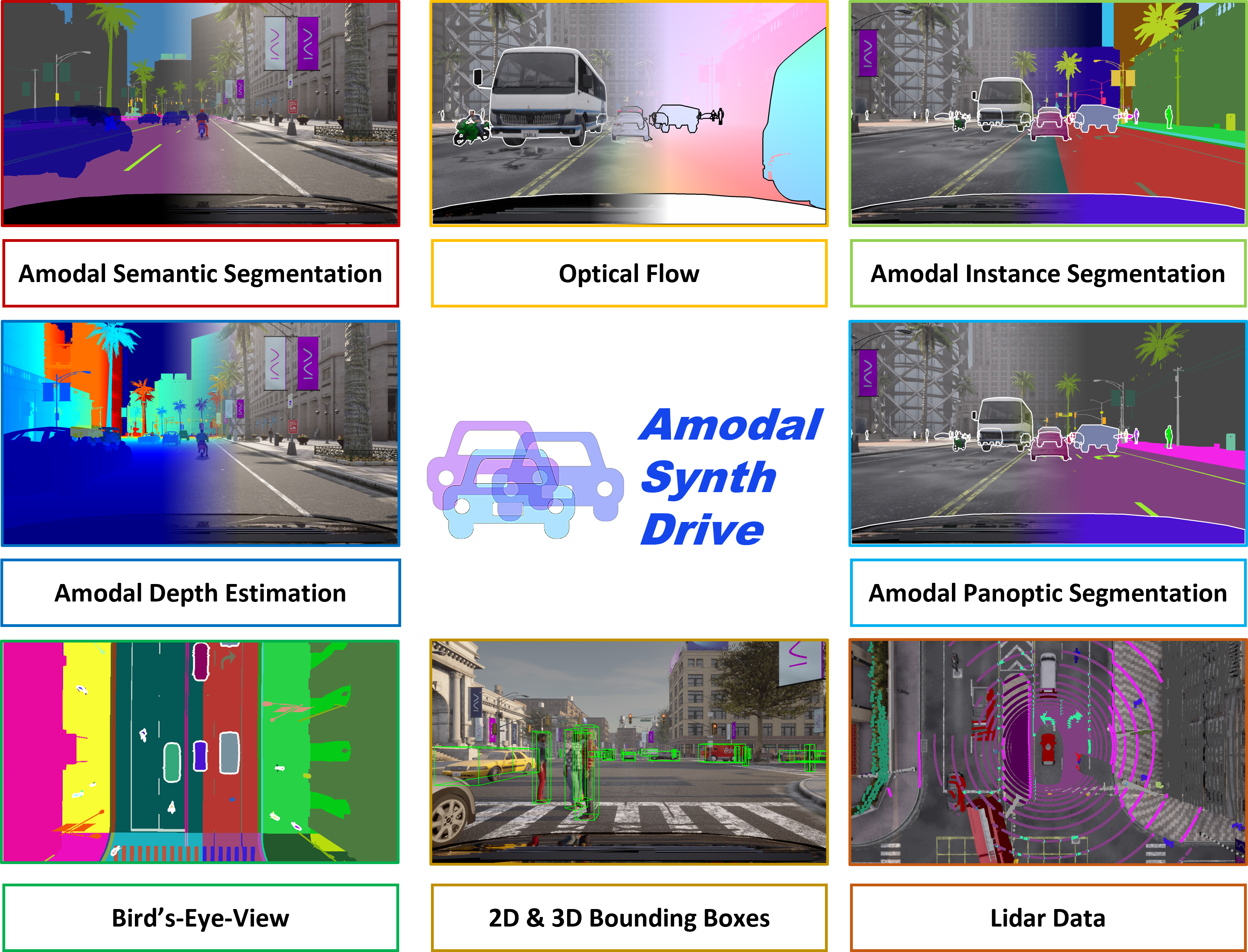}
    \caption{Overview of AmodalSynthDrive consisting of multi-view camera image sequences, 3D bounding boxes, LiDAR data, and odometry. It provides annotations for amodal panoptic segmentation, amodal instance segmentation, amodal semantic segmentation, and amodal depth estimation. It also provides supplementary visual cues such as optical flow and bird's-eye-view maps.}
    \label{fig:teaser}
    \vspace{-0.5cm}
\end{figure}

In this work, we present AmodalSynthDrive, the first large-scale synthetic amodal perception dataset for urban autonomous driving. This dataset consists of scenes captured under varied lighting and weather conditions from multiple cameras, including four surround-view cameras, $360$° LiDAR, and odometry. The dataset offers a range of visual cues such as optical flow, depth, and bird's-eye-view maps, along with ground truth labels for several amodal perception tasks. We also provide annotations for traditional modal perception tasks, including 3D object detection, motion segmentation, and panoptic tracking. \figref{fig:teaser} presents an illustration of our dataset's multifaceted attributes. We believe that AmodalSynthDrive will significantly contribute to the advancement of research in amodal perception in several ways. The dataset facilitates the development of both standalone approaches, which rely on a single data source or modality, and integrated approaches that incorporate multiple data sources~\cite{valada2016convoluted} and visual cues such as bird's-eye-view maps~\cite{gosala2023skyeye} or optical flow, thereby improving robustness. It also enables formulating and evaluating the viability of more complex tasks within the domain of amodal perception.

The AmodalSynthDrive dataset establishes benchmarks for three fundamental amodal perception tasks, namely, amodal semantic segmentation~\cite{breitenstein2022amodal}, amodal instance segmentation~\cite{li2016amodal}, and amodal panoptic segmentation~\cite{mohan2022amodal}. We evaluate existing state-of-the-art approaches on our dataset to demonstrate the challenges and facilitate comparison. Moreover, our work extends beyond the existing tasks by introducing the novel amodal depth estimation task along with suitable evaluation metrics. A primary challenge in depth estimation is occlusion. The inability to accurately express the occlusion relationships between objects results in partial and inaccurate depth estimations, especially around occlusion boundaries~\cite{mertan2022single}. The amodal depth estimation task as illustrated in \figref{fig:tasks}(d) addresses this challenge at the task level by estimating the distance from the camera to all the parts of a scene, encompassing visible amorphous background \textit{stuff} regions and visible regions of \textit{thing} objects as well as occluded regions of \textit{thing} objects. This explicit notion of occlusion aids in better understanding the spatial arrangement and interaction among objects within a scene. Thus, amodal depth estimates provide a more holistic understanding of the scene, including the size and the complete position of all objects without any need for additional data. This comprehensive insight is particularly important in dynamic scenarios where occlusions can change rapidly due to the movement of occluding objects, the ego-vehicle, or both. Consequently, depth estimates of occluded object regions can be leveraged in image-based downstream autonomous driving tasks such as bird’s-eye-view mapping~\cite{ma2022vision}, path planning, and collision avoidance, which are adversely affected by occlusions.\looseness=-1

To summarize, the contributions of this work include:
\begin{itemize}[noitemsep, topsep=-5px]
    \item AmodalSynthDrive, a comprehensive amodal perception dataset for autonomous driving, with diverse data sources.
    \item Benchmarks for fundamental amodal perception tasks, namely, amodal semantic segmentation, amodal instance segmentation, and amodal panoptic segmentation.
    \item The novel amodal depth estimation task aimed at fostering enhanced spatial comprehension. We demonstrate the feasibility of this new task using several novel baselines.
\end{itemize}

\section{Related Work}
\label{sec:related_work}

{\parskip=0pt
 \noindent\textit{Amodal Perception Tasks}: AmodalSynthDrive includes benchmarks for three existing fundamental tasks in amodal perception: amodal panoptic segmentation (APS)~\cite{mohan2022amodal}, amodal instance segmentation~\cite{li2016amodal} (AIS), and amodal semantic segmentation~\cite{purkait2019seeing} (ASS).
The objective of APS~\cite{mohan2022amodal} is to predict the pixel-wise semantic segmentation labels of the visible amorphous regions of \textit{stuff} classes (e.g., road, vegetation, sky, etc.), and the instance segmentation labels of both the visible and occluded countable object regions of \textit{thing} classes (e.g., cars, trucks, pedestrians, etc.). This task allows the prediction of multiple classes and instance-IDs at the pixel level while restricting each segment of a \textit{thing} class to a single instance-ID at the object level. Two existing approaches tackle this task: APSNet~\cite{mohan2022amodal}, a top-down method, and PAPS~\cite{mohan2022perceiving}, a bottom-up technique. APSNet explicitly models the relationship between the occluders and occludes, using three mask heads tailored for segmenting distinct regions and a transformation block for encapsulating the underlying relationships before computing the final outputs. Conversely, PAPS assigns amodal masks to layers based on the relative occlusion order and employs amodal instance regression on each layer while learning background semantics.}

AIS~\cite{li2016amodal} aims at predicting only \textit{thing} category instances along with both visible/modal and amodal (visible + occluded) masks. Similar to instance segmentation, it allows multiple detections of the same object instance. A variety of methods have been proposed for this task, among which ORCNN~\cite{follmann2019learning}, ASN~\cite{qi2019amodal}, BCNet~\cite{ke2021deep}, WALT~\cite{reddy2022walt} and AISFormer~\cite{tran2022aisformer} demonstrate state-of-the-art results. ORCNN~\cite{follmann2019learning} adds an occlusion head as the difference between modal and amodal mask heads whereas ASN~\cite{qi2019amodal} uses an innovative approach that involves an occlusion classification branch for global feature modeling and a multi-level coding block for feature propagation to distinct mask prediction heads. BCNet~\cite{ke2021deep} takes an alternative approach, striving to separate boundaries of occluding and partially occluded objects through the use of two overlapping GCN layers. WALT~\cite{reddy2022walt} focuses on explicitly learning occluder, amodal, and occlusion features.  AISFormer~\cite{tran2022aisformer} distinguishes itself by effectively modeling complex relationships between various masks within an object's regions of interest, viewing these as learnable queries. Lastly, there are two interpretations of ASS in the existing literature. Zhu~\textit{et~al.} describe it as predicting the amodal mask for objects, irrespective of their class. On the other hand, in \cite{purkait2019seeing, breitenstein2022amodal}, ASS is defined as predicting pixel-wise semantic labels for visible and occluded regions of \textit{stuff} and \textit{thing} categories, with no differentiation between identical \textit{thing} category objects. Hence, this task overlooks occlusion within the same class category. For our benchmark, we adhere to the latter interpretation of ASS~\cite{breitenstein2022amodal}. To tackle this task, Breitenstein~\textit{et~al.}~\cite{breitenstein2022amodal} adapt the standard cross-entropy loss to a semantic group setting with one background and multiple foreground object groups. Subsequently, the same authors~\cite{breitenstein2023joint} propose joint training with both modal and amodal semantic segmentation labels. While existing amodal perception tasks provide rich scene understanding, the novel amodal depth estimation offers unique insights into the occluded space dynamics, introducing a critical depth perspective that further enriches holistic scene interpretation.

{\parskip=3pt
\noindent\textit{Datasets}: Typically, annotations in real images are created by humans, which can result in a limited quantity of annotated data~\cite{hurtado2022semantic}. However, the emergence of simulators has addressed this issue by providing large-scale datasets with synthetic data that are meticulously annotated. These datasets are readily accessible to the public~\cite{mohan2023syn} and cater to various computer vision tasks. Synthetic datasets, particularly those generated from open and free simulators, offer a cost-effective alternative to real image datasets~\myworries{such as ACDC~\cite{sakaridis2021acdc} which encompasses four adverse weather conditions.} Several recent datasets~\cite{sekkat2020omniscape, sekkat2022synwoodscape} use the CARLA simulator~\cite{carlasimulator} and GTA V as a source of data. \myworries{The DeLiVER~\cite{zhang2023delivering} dataset gathered in CARLA, offers benchmarks for modal semantic segmentation and SHIFT~\cite{sun2022shift} addresses both discrete and continuous domain shifts for modal perception tasks.} More particularly in~\cite{hu2019sail}, the authors propose SAIL-VOS, a dataset that contains video sequences with amodal instance segmentation labels extracted from the GTA~V game by interacting with the game engine and systematically alternating the visibility of objects to retrieve information from the occluded regions. Although it is challenging to annotate amodal ground truth labels in real-world images, recent work~\cite{qi2019amodal, breitenstein2022amodal, mohan2022amodal, diaz2022ithaca365} has proposed solutions primarily based on semi-automatic annotation techniques. The WALT dataset~\cite{reddy2022walt} proposed a method for automatic occlusion supervision using time-lapse imagery from stationary webcams observing street intersections. \tabref{tab:SummaryTable} compares the properties of existing amodal perception datasets with the proposed AmodalSynthDrive.}

\begin{table*}
\scriptsize
\centering
\caption{Summary of various amodal perception datasets for autonomous driving.}
\begin{tabular}{@{}l|c|c|c|c|c|c|c@{}}
\toprule
                      & KINS~\cite{qi2019amodal}
                      & Amodal Cityscapes~\cite{breitenstein2022amodal}
                      & WALT~\cite{reddy2022walt}
                      & Ithaca365~\cite{diaz2022ithaca365}
                      & KITTI360-APS~\cite{mohan2022amodal}
                      & BDD100K-APS~\cite{mohan2022amodal}
                      & AmodalSynthDrive (Ours) \\ \midrule
                      
Synthetic or Real     & Real & Real & Real & Real                   & Real & Real & Synthetic \\ \midrule
Depth Map             & \no  & \no  & \no  & \no                    & \no  & \no  & \yes \\ \midrule
3D Bounding Boxes     & \no  & \no  & \no  & \yes                   & \no  & \no  & \yes \\ \midrule
Amodal Semantic Seg.  & \no  & \yes & \no  & \yes (only class Road) & \no  & \no  & \yes \\ \midrule
Amodal Instance Seg.  & \yes & \no  & \yes & \yes                   & \no  & \no  & \yes \\ \midrule
Amodal Panoptic Seg.  & \no  & \no  & \no  & \no                    & \yes & \yes & \yes \\ \midrule
Amodal Motion Seg.    & \no  & \no  & \no  & \no                    & \no  & \no  & \yes \\ \midrule
Amodal Depth Map      & \no  & \no  & \no  & \no                    & \no  & \no  & \yes \\ \midrule
Optical Flow          & \no  & \no  & \no  & \no                    & \no  & \no  & \yes \\ \midrule
LiDAR                 & \no  & \no  & \no  & \yes                   & \no  & \no  & \yes \\ \midrule
IMU/GNSS              & \no  & \no  & \no  & \yes                   & \no  & \no  & \yes \\ 
\bottomrule
\end{tabular}
\vspace{-0.2cm}

\label{tab:SummaryTable}
\end{table*}

\section{AmodalSynthDrive Dataset}
\label{sec:dataset}
In this section, we provide details about the AmodalSynthDrive dataset, including its collection process in~\secref{subsec:creation}, the dataset structure in~\secref{subsec:structure}, benchmarking tasks in~\secref{subsec:benchmark}, and dataset statistics in~\secref{subsec:stats}.

\begin{figure*}
    \centering
        \includegraphics[width=0.85\linewidth]{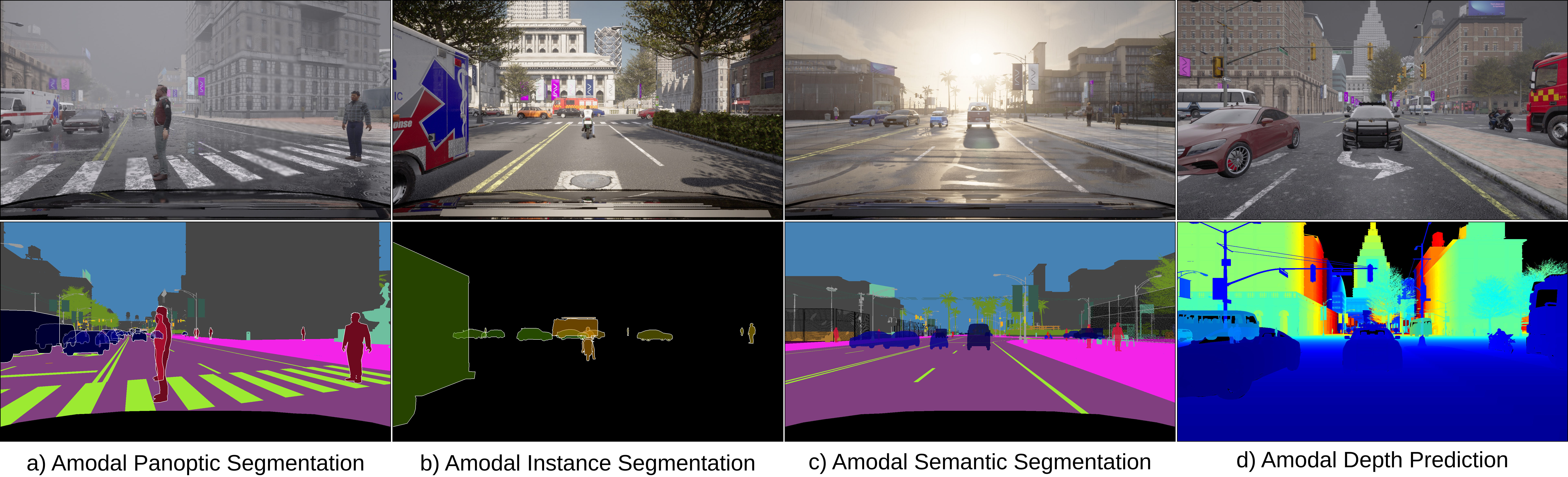}
    \caption{Illustration of different amodal perception tasks in AmodalSynthDrive. For optimal viewing, we recommend magnifying the figure by four times.}
    \label{fig:tasks}
    \vspace{-0.2cm}
\end{figure*}

\subsection{Dataset Collection}
\label{subsec:creation}

We create the AmodalSynthDrive dataset using the CARLA simulator~\cite{carlasimulator} (under MIT license). It enables populating a road scene with different vehicles and pedestrians, setting the world parameters such as weather conditions through a server-client interface. To compute the amodal ground truth labels, we need to access information from the occluded region, which is not visible on the screen and is not directly extracted from the simulator. We took advantage of the communication interface to systematically alternate the visibility of objects in the scene to enable the perception of each object in its entirety. In the generation process, the first step involves the initialization of the simulation with the world parameters such as the weather conditions. Afterward, we create a scenario randomly using CARLA's autopilot features that allow actors to drive and move autonomously as real road traffic. In parallel, we save all the necessary information that allows the re-rendering of said scenario. Once the scenario has been created, we render all the visible objects separately without including the background one by one and frame by frame. For each object being rendered, we extract the ground truth labels using the five cameras: front; back; left; right; and the bird's-eye-view; in a manner to maintain a unique instance ID in all the annotations. This is essential for tracking object instances across the different camera views.

\subsection{Dataset Structure}
\label{subsec:structure}

From the CARLA simulation environment, we have compiled a total of 60,000 images, each with a resolution of $1080 \times 1920$. Our dataset comprises 150 distinct video sequences that cater to a wide spectrum of situations including different weather conditions, illumination conditions, scenes, and traffic scenarios, as shown in~\figref{fig:tasks}. We accumulated data from a variety of sources including multiple cameras, specifically four surround-view cameras, in addition to LiDAR and odometry data. The AmodalSynthDrive dataset provides an assortment of visual cues such as optical flow, depth, and bird's-eye-view maps. As for the annotations, modal annotations are provided for all surround-view cameras and bird's-eye-view maps. While the amodal annotations are provided solely for the surround-view cameras, encompassing semantic, instance, and panoptic segmentation tasks. AmodalSynthDrive consists of 18 distinct semantic classes. We provide instance annotations for 7 of the semantic classes. The semantic classes in our dataset are detailed in~\figref{fig:inst_num}. In this figure, the semantic classes linked with the blue bar are categorized as \textit{stuff}, while those associated with the orange bar belong to the \textit{thing} category. To further enrich the dataset, we go beyond modal depth ground truth and include the amodal depth of occluded objects. In addition, we furnish ground truths for 3D bounding boxes, motion segmentation, and panoptic tracking for comprehensive analysis and application. Lastly, our dataset encompasses a variety of weather conditions, including clear, cloudy, dusty, and rainy with varying intensities (hard, medium, and soft), as well as different times of the day, namely Noon and Sunset.

Our dataset is partitioned based on geographical considerations into training, validation, and test sets. The training set encompasses 105 video sequences, totaling 42,000 images. The validation subset comprises 15 video sequences (6,000 images), and the test subset holds 30 video sequences, representing 12,000 images. This specific organization is maintained across all the benchmarking tasks in our dataset. It is important to note that the annotations for the test set will not be made publicly available, but will rather be used for the online benchmark evaluation server. 
Consequently, the validation subset is recommended for model parameter tuning or selection prior to performing evaluations on the test server.\looseness=-1

\begin{figure*}
    \centering
        \includegraphics[width=0.7\linewidth]{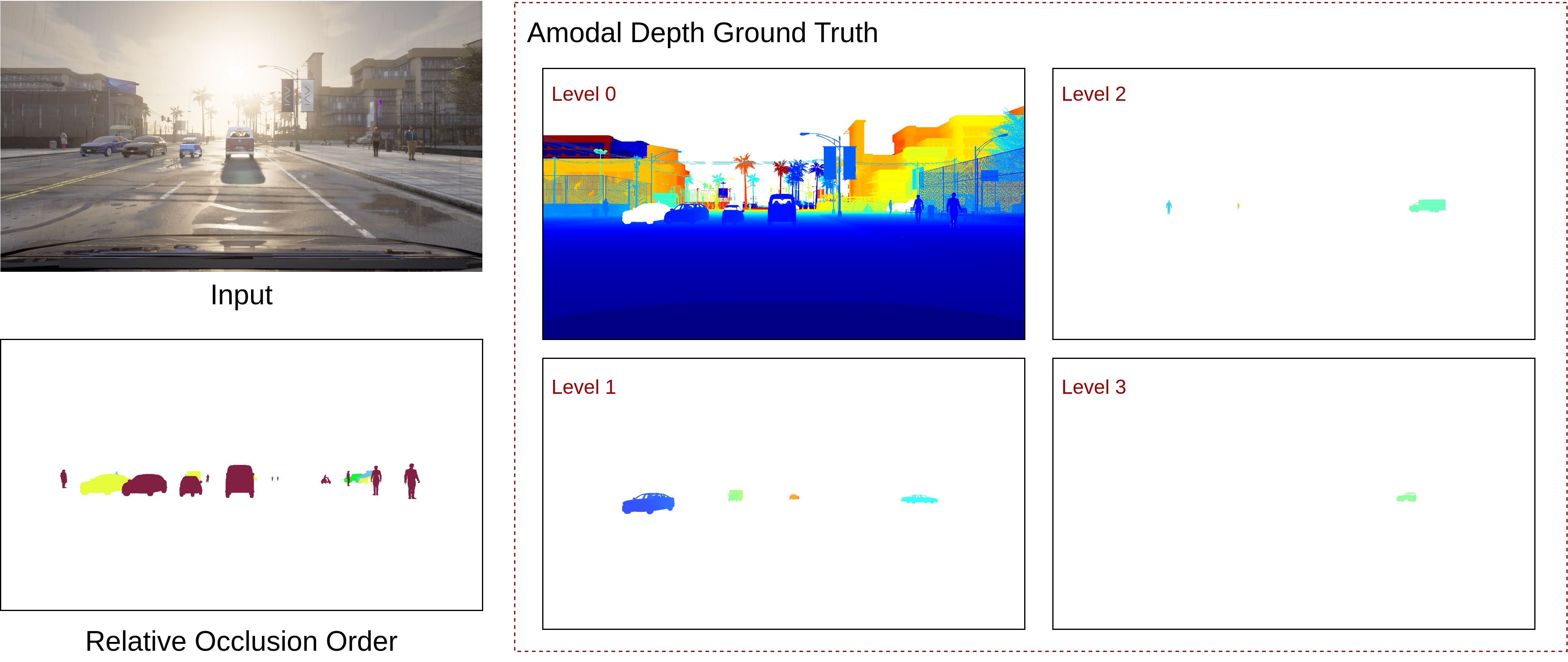}
    \caption{Illustration of amodal depth estimation ground truth, disassembling the scene according to relative occlusion order. Objects are color-coded by level: red for level 0 (excluding amorphous regions for simplicity), yellow for level 1, blue for level 2, and green for level 3.}
    \label{fig:amdepth}
    \vspace{-0.2cm}
\end{figure*}

\subsection{Benchmarking Tasks}
\label{subsec:benchmark}

In this section, we study the task of amodal depth estimation in detail, offering a formal definition, outlining its evaluation metrics, and elaborating on its unique challenges. Subsequently, we focus on describing the nuances of our benchmarking tasks - amodal panoptic segmentation, amodal instance segmentation, and amodal semantic segmentation, in the context of our dataset. For the definition of each of these tasks, please refer to~\secref{sec:related_work} and, for deeper insights, refer to the associated papers.

\subsubsection{Amodal Depth Estimation}
\label{sec:ade}

The objective of amodal depth estimation is to ascertain the distance from the camera to every component of a scene. This includes not only the visible unstructured amorphous \textit{stuff} regions and visible \textit{thing} object regions but also parts of \textit{thing} objects that are occluded. We limit depth estimation for amorphous regions to directly visible portions only, as these regions lack the structure or identifiable features necessary for amodal prediction. Consequently, this task entails creating a complete depth profile that includes an understanding of the depth of unseen areas based on the depth and characteristics of the visible parts of the scene. In order to facilitate multiple depth values for a single pixel, which could relate to either visible or hidden regions of the scene's objects, we leverage the principle of relative occlusion order within a scene.
The relative occlusion order is a tiered approach to classify scene elements based on their visibility and occlusion potential. Divided into $N$ levels, each level encapsulates a unique occlusion degree:
\begin{itemize}[topsep=0pt]
\item Level 0: All amorphous regions and unoccluded objects.
\item Level 1 to Level N-1: Each subsequent level comprises objects occluded by at least one object from the preceding level and any objects from the levels before that. Notably, objects within the same level do not occlude each other, ensuring the distinctness of amodal masks.
\end{itemize}

Through this ordering scheme, a pixel can potentially possess multiple depth values, each corresponding to a different occlusion level it is associated with. This method not only contributes to a more comprehensive understanding of the scene but also systematically elucidates the depth relationships amongst various elements, thereby facilitating the estimation of depth in both visible and occluded regions. \myworries{We set the maximum relative occlusion order N to 8, as this represents the highest possible relative occlusion order in our dataset.} \figref{fig:amdepth} illustrates the relative occlusion order for an example scene and the corresponding depth ground truth for each order. 

{\parskip=3pt
 \noindent\textit{Task Definition}: For the given input $I$, the task aims to generate a series of depth maps $\{D_0, D_1,\dots,D_{N-1}\}$, each corresponding to a specific occlusion level $(n)$ within the scene. Here, each depth map $D_n$ is identical in dimensions to the input, with each pixel value denoting the depth at that point. Functionally, this can be denoted as
\begin{equation}
    f(I) \to \{D_0, D_1, ..., D_{N-1}\}.
\end{equation}
 This task extends beyond the typical depth estimation, requiring comprehension and prediction of both visible and occluded sections, facilitating a comprehensive understanding of the scene's spatial structure and relationships.
 }

{\parskip=3pt
 \noindent\textit{Evaluation Metric}: To evaluate the amodal depth estimation task, we use the Root Mean Square Error (RMSE), a commonly used metric for depth estimation. We denote the predicted depth map for occlusion level $n$ as $D_n^p$ and the ground truth depth map for occlusion level $n$ as $D_n^g$. We then define the Root Mean Square Error (RMSE) for each occlusion level $n$ as
 \begin{equation}
RMSE_n = \sqrt[2]{\frac{1}{N_n} \sum_{i,j} (D_{n,i,j}^p - D_{n,i,j}^g)^2},
\end{equation}
where the summation is over all pixels $(i,j)$ in the depth map, and $N_n$ represents the total number of pixels.
Having calculated the individual RMSE values for each occlusion level, we compute the weighted mean over all $N$ occlusion levels to obtain the final evaluation metric called Amodal Depth Error ($ADErr$) computed as
\begin{equation}
ADErr = \sum_{n=0}^{N-1} w_n \cdot RMSE_n.
\end{equation}
where $w_n$ represents the weight assigned to layer $n$, defined as the ratio of the total number of occurrences of layer $n$ to the total number of occurrences of all layers across the dataset. We introduce $w_n$ to counter the imbalance in the distribution of occlusion layers, as evidenced in~\figref{fig:occ_asd2}. The resulting evaluation metric $ADErr$, computes the depth estimation accuracy across all occlusion levels, offering a holistic performance measure that encompasses both visible and occluded regions.}

{\parskip=3pt
\noindent\textit{Challenges}: Amodal depth estimation presents several complexities. Primarily, occlusion handling necessitates extrapolating from available data to predict the depth of occluded regions of objects. Furthermore, discerning the relative occlusion order is essential for establishing an accurate depth hierarchy, requiring the model to comprehend depth relationships among objects. The task complexity is also amplified by the need to generate multiple depth maps, each corresponding to a distinct occlusion level. The model must efficiently distinguish these levels and yield precise estimations for each, given the varying occlusion degrees. Collectively, these complexities underline the need for advanced strategies that exploit the multi-camera setup, supplementary visual cues, and the temporal dimension of our dataset for reliable amodal depth estimation.}

\subsubsection{Amodal Perception Tasks}

{\parskip=0pt
\noindent\textit{Amodal Panoptic Segmentation}: In amodal panoptic segmentation~\cite{mohan2022amodal}, we account for 17 semantic classes. Among these, 11 are classified as \textit{stuff} classes, while the remaining 6 fall under \textit{thing} classes. A class is designated as \textit{stuff} if its regions are amorphous or it is incapable of movement at any given point in time. The \textit{stuff} classes comprise road, sidewalk, building, wall, fence, pole, traffic light, traffic sign, vegetation, terrain, and sky. Additionally, the 6 \textit{thing} classes consist of a person, car, bike, motorcycle, truck, and bus. For evaluating this task, we employ the amodal panoptic quality and amodal parsing coverage metrics as defined in~\cite{mohan2022amodal}. As baselines for this task,  we train and benchmark two existing methods for amodal panoptic segmentation, namely APSNet~\cite{mohan2022amodal} and PAPS~\cite{mohan2022perceiving}, as well as the AmodalEfficientPS, a naive baseline proposed in~\cite{mohan2022amodal}. An illustration of the ground truth for this task is presented in~\figref{fig:tasks}a.\looseness=-1

{\parskip=3pt
\noindent\textit{Amodal Instance Segmentation}:
The amodal instance segmentation~\cite{li2016amodal} task in our dataset focuses on 6 semantic classes, namely person, car, bike, motorcycle, truck, and bus. We use the average precision ($AP$) of modal ($AP_{modal}$) and amodal ($AP_{amodal}$) masks~\cite{li2016amodal} metric for evaluations. Furthermore, we train and benchmark the three leading state-of-the-art methods on AmodalSynthDrive: ASN~\cite{qi2019amodal}, BCNet~\cite{ke2021deep}, and AISFormer~\cite{tran2022aisformer}. \figref{fig:tasks}b shows the ground truth for this task.}

{\parskip=3pt
\noindent\textit{Amodal Semantic Segmentation}:
For amodal semantic segmentation~\cite{breitenstein2022amodal}, our dataset focuses on the evaluation of 17 semantic classes, which can be broadly grouped into static and dynamic categories. Occlusions within the static group are not considered for this task, whereas occlusions between the static and dynamic groups or within the dynamic group itself are evaluated. The static group, similar to the \textit{stuff} classes in amodal panoptic segmentation, includes road, sidewalk, building, wall, fence, pole, traffic light, traffic sign, vegetation, terrain, and sky. The dynamic \textit{thing} classes consist of person, car, bike, motorcycle, truck, and bus. \figref{fig:tasks}c presents an example of this task. Please note that in the ground truth, the occlusion between the two cars on the left is not taken into account. However, the occluded areas behind one of the cars, particularly for the truck, do need to be segmented. Similarly, the static group regions occluded by the truck also need to be appropriately segmented. We use the $mIoU\textsuperscript{total}$ metric for evaluating this task, as outlined in~\cite{breitenstein2022amodal}. Additionally, we train and benchmark two state-of-the-art methods amERFNet~\cite{breitenstein2022amodal} and Y-ERFNet~\cite{breitenstein2023joint}.}

\begin{figure}
    \centering
    \begin{subfigure}[b]{0.49\linewidth}
        \includegraphics[width=\linewidth]{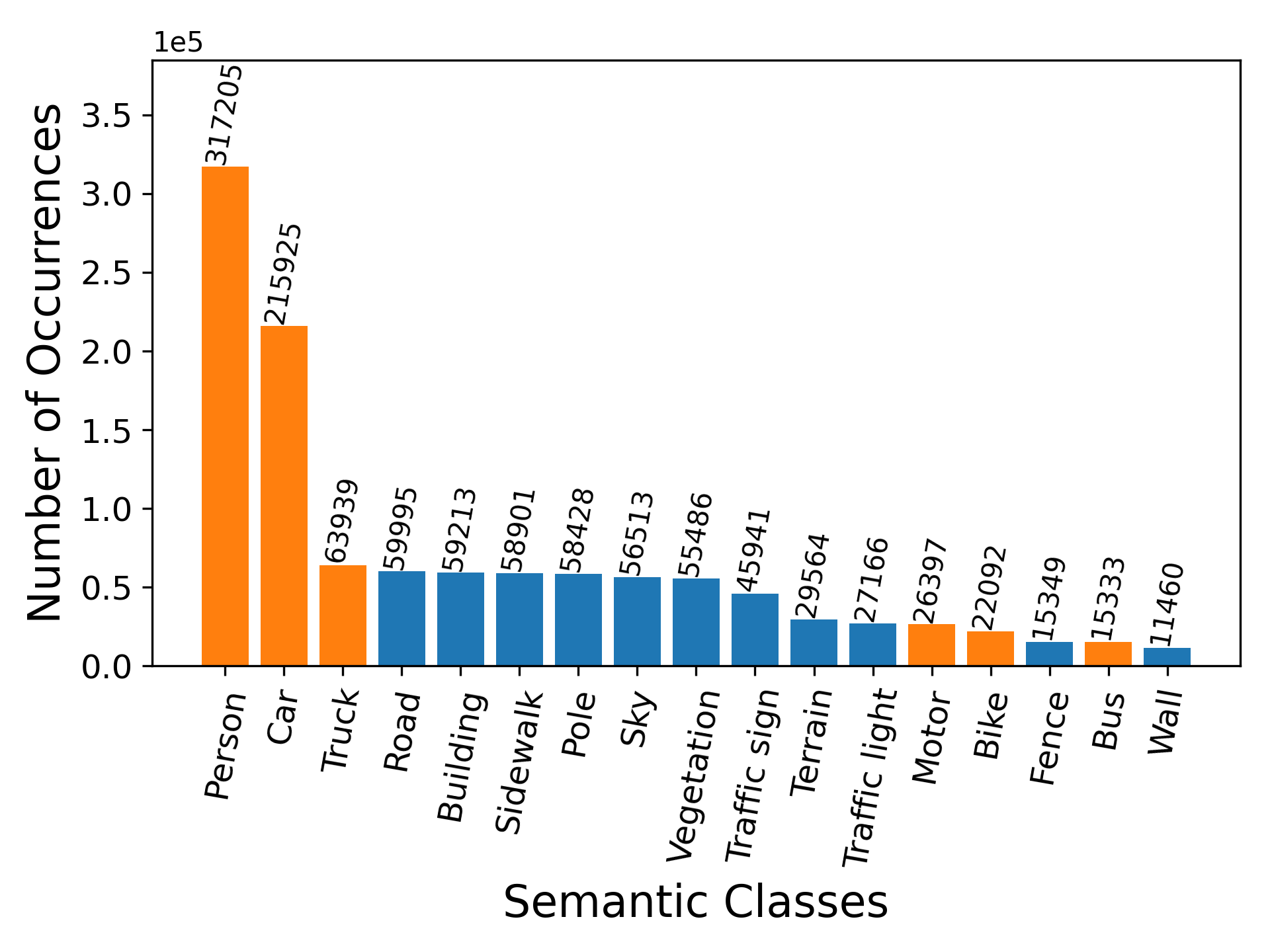}
        \subcaption{Number of segments per semantic class.}
        \label{fig:inst_num}
    \end{subfigure}
        \begin{subfigure}[b]{0.49\linewidth}
        \includegraphics[width=\linewidth]{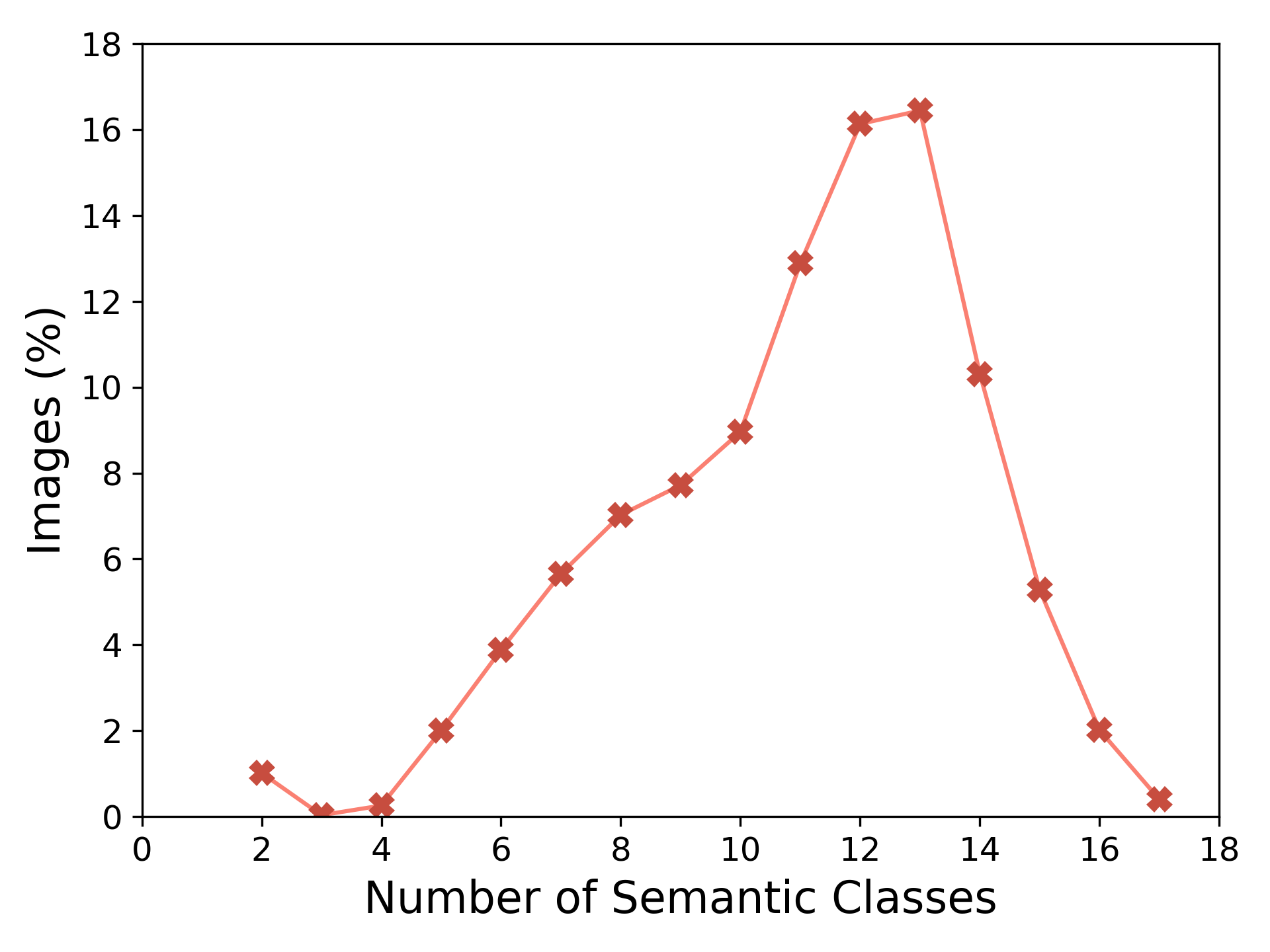}
        \subcaption{Distribution of images by semantic class count.}
        \label{fig:occ_asd}
    \end{subfigure}
    \begin{subfigure}[b]{0.49\linewidth}
        \includegraphics[width=\linewidth]{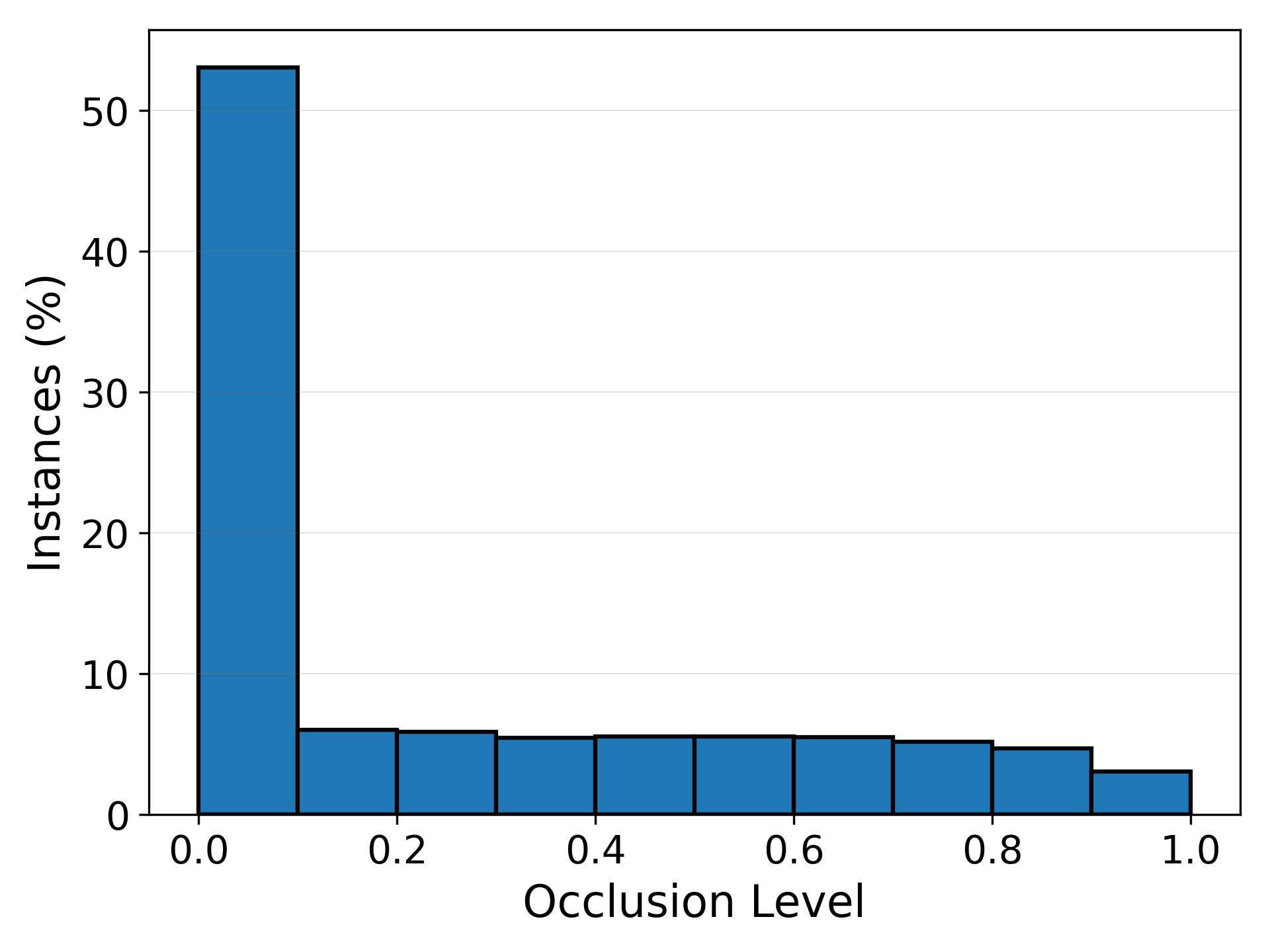}
        \subcaption{Distribution of instances by occlusion level} 
        \label{fig:occ_asd1}
    \end{subfigure}
    \begin{subfigure}[b]{0.49\linewidth}
        \includegraphics[width=\linewidth]{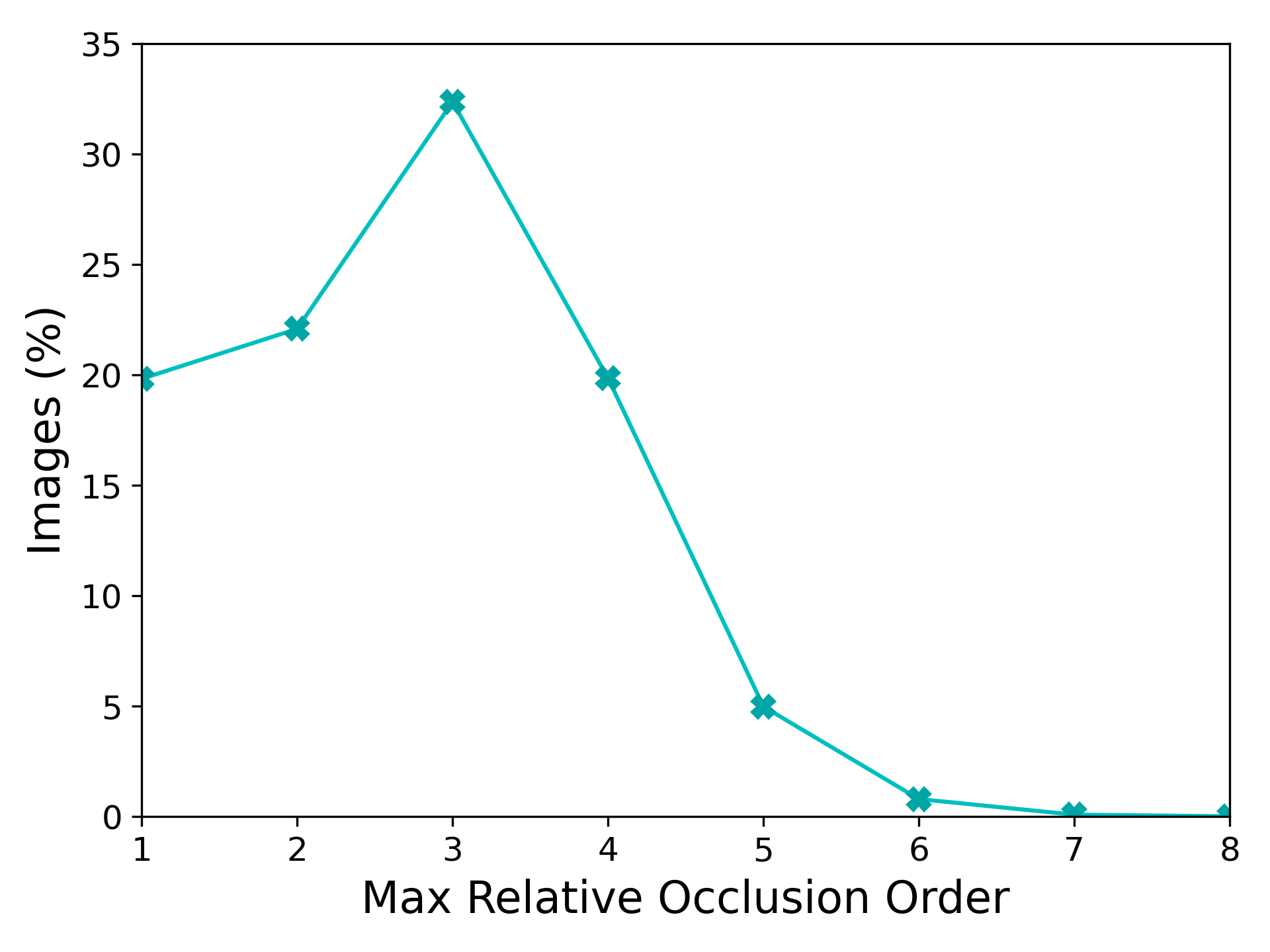}
        \subcaption{Distribution of images by occlusion order}
        \label{fig:occ_asd2}
    \end{subfigure}
    \caption{Dataset statistics of AmodalSynthDrive.}
    \label{fig:occ}
    \vspace{-0.5cm}
\end{figure}

\begin{table}
\scriptsize
\centering
\renewcommand\arraystretch{1}
\renewcommand{\tabcolsep}{2mm}
\caption{Comparison of shape statistics between modal and amodal segments in our introduced AmodalSynthDrive dataset, along with KINS, KITTI-360-APS, and BDD100K-APS datasets.}
\begin{tabular}{l|cc}
\toprule
 & Simplicity\textsubscript{Modal} & Simplicity\textsubscript{Amodal} \\
\midrule
KINS~\cite{qi2019amodal}  & 0.709   & 0.830  \\
KITTI-360-APS~\cite{mohan2022amodal} & 0.778   & 0.884   \\
BDD100K-APS~\cite{mohan2022amodal}  & 0.697   & 0.821   \\
AmodalSynthDrive (Ours) & \textbf{0.585}   & \textbf{0.633}  \\
\bottomrule
\end{tabular}
\vspace{-0.2cm}

\label{tab:shape}
    \vspace{-0.2cm}
\end{table}

\subsection{Dataset Statistics and Analysis}
\label{subsec:stats}
\figref{fig:occ} presents a comprehensive statistical overview of our dataset. \figref{fig:inst_num} presents the distribution of object segments divided by semantic class, with \textit{stuff} classes denoted by blue bars and \textit{thing} classes, those with distinct instances, represented by orange bars. Our dataset stands out for its substantial proportion of person instances. The complexity of our dataset is further highlighted in \figref{fig:occ_asd} which shows the image distribution by semantic class count with a notable peak between 8 to 15 classes. 
\figref{fig:occ_asd1} illustrates the distribution of instances according to their occlusion level, providing insights into the wide occlusion spread inherent in our dataset. Following,~\figref{fig:occ_asd2} presents the distribution of images based on the maximum relative occlusion order, indicating that while our data can accommodate up to 8 relative occlusion orders, most scenes feature between 2 to 4 levels. Moreover, our dataset's mask shape characteristics are compared with other amodal autonomous driving datasets in \tabref{tab:shape}, quantified by simplicity~\cite{zhu2017semantic}. Due to the high precision of masks in synthetic datasets compared to real-world annotations, our dataset exhibits a lower simplicity for both modal and amodal masks, making it a challenging benchmark for amodal perception.  

\section{Experiments}
\label{sec:experiments}

In this section, we present benchmarking results on the test set of AmodalSynthDrive for all the amodal perception tasks as described in \secref{subsec:benchmark}. Subsequently, we discuss the impact of training methods with our dataset in conjunction with real-world datasets specific to amodal instance segmentation and amodal panoptic segmentation. For details on the training protocols, refer to the supplementary materials.

\begin{table*}
\caption{Benchmarking results of different amodal perception tasks on the test set of AmodalSynthDrive.}
  \begin{subtable}{0.35\linewidth}
  \scriptsize
    \centering
    
    \caption{Amodal Panoptic Segmentation}
    \begin{tabular}{l|cc}
    \toprule
      Method & APQ & APC \\
      \midrule
      Amodal-EfficientPS~\cite{mohan2022amodal} & $45.7$ &$60.7$\\
      APSNet~\cite{mohan2022amodal} & $48.6$ &$62.5$\\
      PAPS~\cite{mohan2022perceiving} & $\mathbf{50.5}$ & $\mathbf{64.1}$ \\
      \bottomrule
    \end{tabular}
    
    \label{tab:aps}
  \end{subtable}
  \begin{subtable}{.35\linewidth}
  \scriptsize  
    \centering
    \caption{Amodal Instance Segmentation}
    \begin{tabular}{l|cc}
      \toprule
      Method & AP$_{amodal}$ & AP$_{modal}$ \\
      \midrule
        ORCNN~\cite{follmann2019learning} & $46.1$ & $43.8$\\
        ASN~\cite{qi2019amodal} & $48.6$ & $45.2$\\
        BCNet~\cite{ke2021deep} &$49.4$ & $46.7$\\
        WALT~\cite{reddy2022walt} & $49.9$ & $47.2$\\
        AISFormer~\cite{tran2022aisformer} & $\mathbf{50.3}$& $\mathbf{47.7}$ \\
        \bottomrule
    \end{tabular}
    
    \label{tab:ais}
  \end{subtable}
  \begin{subtable}{.35\linewidth}
    \scriptsize
    \centering
    \caption{Amodal Semantic Segmentation}
    \begin{tabular}{l|c}
      \toprule
      Method & mIoU$^{total}$  \\
      \midrule
        amERFNet~\cite{breitenstein2022amodal} & $38.6$\\
        Y-ERFNet~\cite{breitenstein2023joint} & $\mathbf{42.4}$\\
        \bottomrule
    \end{tabular}

    \label{tab:ass}
  \end{subtable} 
  \vspace{-0.4cm}
  
\end{table*}

\begin{table}
    \scriptsize
    \centering
    \caption{Benchmarking Results of amodal depth estimation task on the test set of AmodalSynthDrive}
    \begin{tabular}{l|cc}
      \toprule
      Method & ADErr & RMSE$_{vis}$ \\
      \midrule
        Amodal-Dorn & $23.9$ & $10.13$\\
        ADB-DeepLab & $23.2$ & $9.59$\\
        AD-DeepLab & $\mathbf{21.6}$ & $\mathbf{7.93}$\\
        \bottomrule
    \end{tabular}

    \label{tab:ade}
\end{table}

\subsection{Benchmarking of  Amodal Perception Tasks}
We present results for amodal panoptic segmentation (APS), amodal instance segmentation (AIS), and amodal semantic segmentation (ASS) in \tabref{tab:aps}, \tabref{tab:ais}, and \tabref{tab:ass} respectively. We observe that PAPS outperforms others for APS, while AISFormer excels in AIS. The performance scores for these models highlight the complexity of our dataset for these tasks. These top-performing models typically enhance their performance by learning explicit occluding features through either occluder and occluded region masks or by estimating relative occlusion order. Similarly, for ASS, a joint training approach with both modal and amodal semantic segmentation maps yields improved results, as it compels the model to effectively distinguish between visible and occluded areas while providing some contextual estimate of the underlying semantic class. However, considering the diverse weather conditions and detailed occluded region annotations in our dataset, we conjecture that solely depending on explicit occlusion feature encoding may not suffice. A more robust approach for addressing amodal perception tasks would likely require exploiting our dataset's sequential nature, utilizing multi-modality and available visible cues, along with sharper prior encodings.

\subsection{Benchmarking of  Amodal Depth Estimation}

\begin{figure*}
    \centering
        \includegraphics[width=0.75\linewidth]{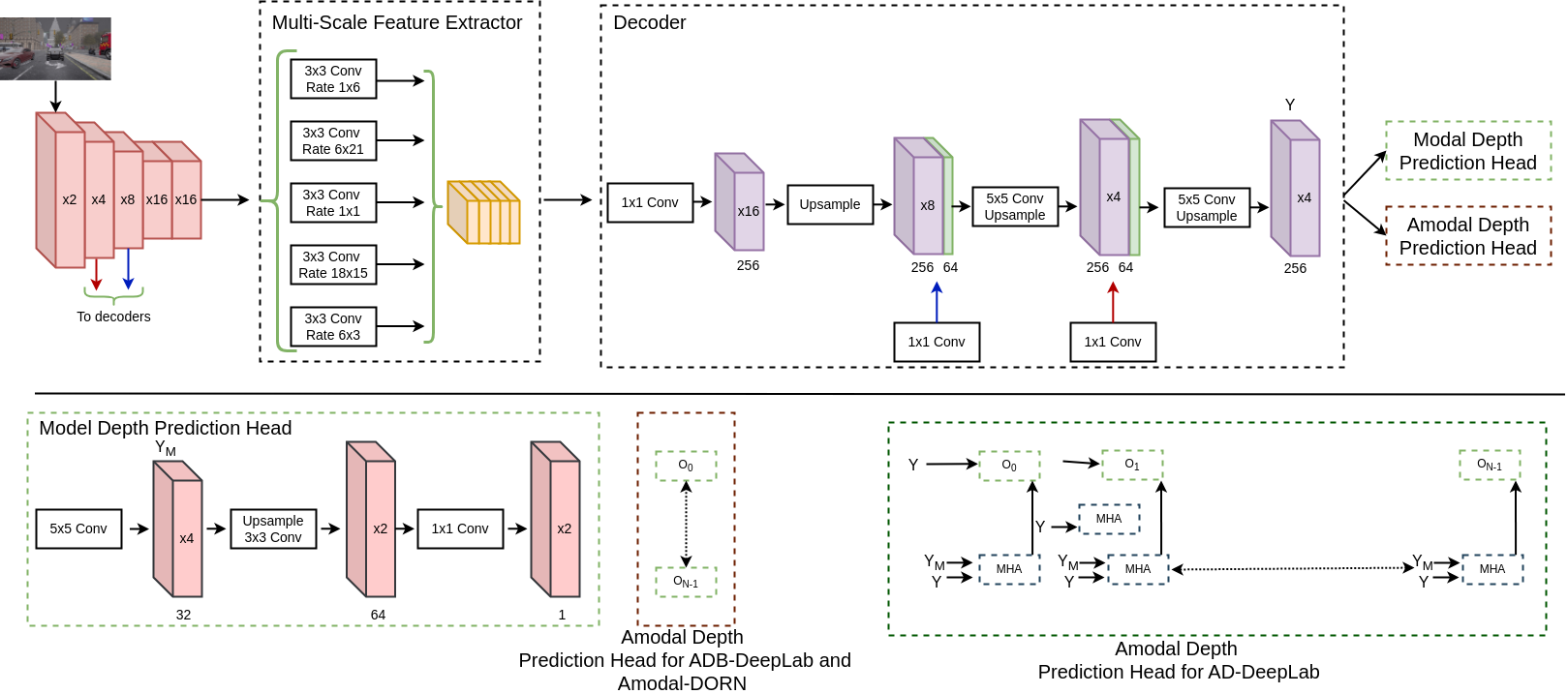}
    \caption{Architectural topology of the amodal depth estimation baseline: ADB-DeepLab, Amodal-DORN and AD-Deeplab. Please note that the boxes enclosed in colored dashes in each of the architecture correspond to the expanded version of the same colored boxes depicted on the bottom.}
    \label{fig:arch}
\end{figure*}

In order to study the feasibility of our proposed amodal depth estimation (ADE) task, we perform experiments using two baselines, employing a network akin to VIP-DeepLab~\cite{qiao2021vip} as our backbone due to its simple architecture design. \figref{fig:arch} illustrates the architectures of these baselines. This network, modified to retain only the depth estimation component, comprises an encoder, a multi-scale context extractor, and a decoder with a modal depth estimation head. In the first model, Amodal Depth Basic-Deeplab (ADB-Deeplab), we enhance the decoder by incorporating eight depth estimation heads, each dedicated to predicting depth maps of the scene at different occlusion levels. Additionally, for each depth estimation head, we incorporate a binary mask predictor to determine whether the depth is valid or invalid. In Amodal-DORN, we employ the spacing-increasing discretization strategy, as mentioned in~\cite{fu2018deep}, to discretize depth and reformulate it as an ordinal regression problem. We report ADErr (\secref{sec:ade}) evaluation metric for amodal depth estimation and RMSE$_{vis}$ for the modal depth in~\tabref{tab:ade}. To generate the depth map of only the visible region, we stack all depth maps, starting from the furthest occlusion level to no occlusion. Each successive layer overwrites the depth values of the previous one at locations where the depth is valid. We hypothesize that the significant disparity between ADE and RMSE$_{vis}$ errors indicates the network's difficulty in learning the delineated multiple depth maps directly. 

For the second model, Amodal Depth-Deeplab (AD-Deeplab), we incorporate multi-headed attention feature propagation blocks across the various depth estimation heads. For each depth estimation head at level $l$, the modal depth head generates multiple key and value embeddings. Similarly, the fused features from up to the $l-1$ layer yield a distinct set of multiple key and value embeddings, tailored to that specific depth layer. The multi-headed attention blocks utilize the amodal features of the $l$th layer as their query. These features are then combined with the attended features from the multi-headed attention block, subsequently processed through a $5\times5$ depthwise separable convolution, and finally fed into the $1\times1$ depth prediction head. This model exhibits an improvement in ADErr and RMSE$_{vis}$. This increase in RMSE$_{vis}$ also implies that advancements in amodal depth estimation through enhanced occlusion reasoning capabilities subsequently lead to improved modal depth estimates. In AD-Deeplab, the inclusion of attention-based feature propagation between different depth estimation heads enhances the flow of information, thereby facilitating more accurate occlusion estimation in the scene.  \figref{fig:qual} presents the qualitative results of AD-DeepLab on our proposed AmodalSynthDrive dataset. The model demonstrates its capability to distinguish between scene element layers and estimate the depth of occluded regions to a certain degree, as seen with the car behind the bus in (a) and the motorbike in (b). However, the qualitative and quantitative results from our baselines exhibit the high complexity of the task and the necessity for introducing explicit occlusion encoding mechanisms while finding the right balance to keep the performance of visible depth estimation. As the research community delves deeper into this novel task, we hope these initial explorations serve as a foundation, guiding the design of more sophisticated models for robust amodal depth estimation in the future. Please refer to the supplementary material for a detailed description of the baseline model architectures.

\begin{table}
\centering
\scriptsize
\caption{Comparison of amodal instance segmentation performance with/without AmodalSynthDrive on KINS.  All scores are in [\%].}
\begin{tabular}
{l|l|cc}
\toprule
Dataset & Method & AP$_{amodal}$ & AP$_{modal}$ \\
\toprule
\multirow{2}{*}{KINS} &BCNet & $32.6$&  $30.4$\\ 
& AISFormer & $33.8$&  $32.1$ \\
\midrule
KINS +  & BCNet & $\mathbf{34.5}$&  $\mathbf{32.2}$ \\ 
AmodalSynthDrive & AISFormer & $\mathbf{35.9}$&  $\mathbf{33.6}$ \\
\bottomrule
\end{tabular}
\vspace{-0.2cm}

\label{tab:amodalInstance}
\end{table}

\begin{table}[t]
\centering
\scriptsize
\caption{Comparison of amodal panoptic segmentation performance with/without AmodalSynthDrive on KITTI-360-APS.  All scores in [\%].}
\begin{tabular}
{l|l|cc}
\toprule
Dataset & Method & APQ & APC \\
\toprule
\multirow{2}{*}{KITTI-360 APS} &Amodal-EfficientPS & $41.1$&  $57.6$\\ 
& APSNet & $42.9$&  $59.0$ \\
\midrule
KITTI-360 APS +  & Amodal-EfficientPS & $\mathbf{43.1}$&  $\mathbf{59.5}$ \\ 
AmodalSynthDrive & APSNet & $\mathbf{45.4}$&  $\mathbf{60.9}$ \\
\bottomrule
\end{tabular}
\vspace{-0.2cm}

\label{tab:amodalPanoptic}
\vspace{-0.3cm}
\end{table}

\begin{figure*}
    \centering
    \includegraphics[width=\linewidth]{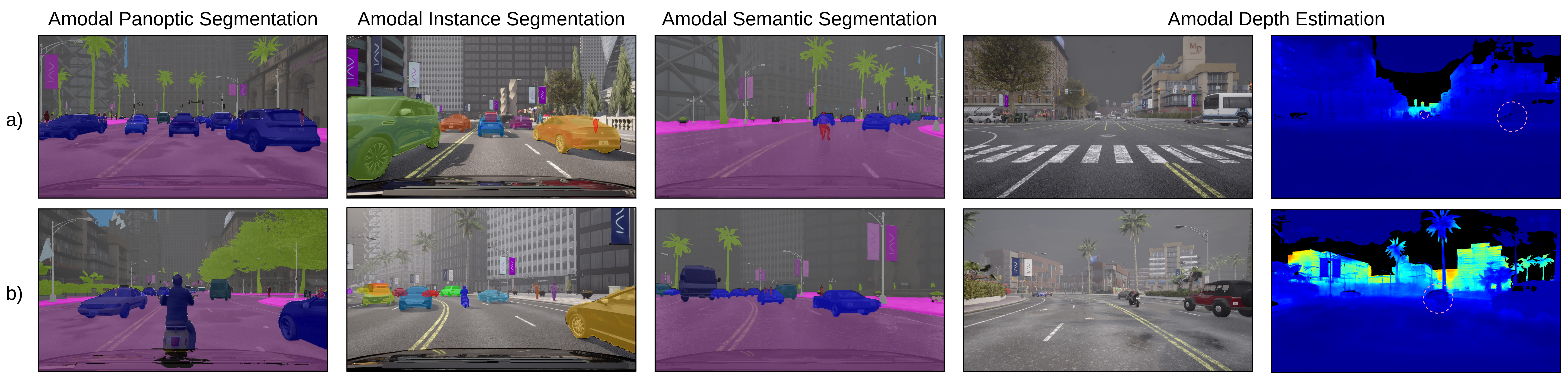}
    
    \caption{Qualitative results of different amodal perception tasks in the AmodalSynthDrive dataset. For amodal depth estimation, we sequentially superimpose its multi-layer predictions. To enhance viewing, magnifying the figure fourfold is recommended.
    }
    \label{fig:qual}
\end{figure*}

\subsection{Transfer Learning Results on Real-World Data}

In this section, we study the utility of AmodalSynthDrive for transfer learning on real-world data for amodal panoptic segmentation. To do so, we initially trained the models on AmodalSynthDrive for amodal instance segmentation and amodal panoptic segmentation. Subsequently, we further train these models on their respective real-world datasets: KINS for amodal instance segmentation and KITTI-360-APS for amodal panoptic segmentation. We train BCNet~\cite{ke2021deep} and AISFormer~\cite{tran2022aisformer} for amodal instance segmentation, with results detailed in~\tabref{tab:amodalInstance}. For amodal panoptic segmentation, we employed Amodal-EfficientPS~\cite{mohan2022perceiving} and APSNet~\cite{mohan2022perceiving}, and the results for these are presented in \tabref{tab:amodalPanoptic}. Our findings show a $1.9-2.1\%$ increase in AP$_{amodal}$ score for amodal instance segmentation and about $2-2.5\%$ improvement in the APQ score for amodal panoptic segmentation. These results substantiate the claim that pre-training on AmodalSynthDrive improves the performance on real-world amodal perception datasets. Furthermore, the performance achieved by first pretraining on synthetic data and then finetuning on real-world data, without domain adaptation, is consistent with the improvement reported in other established synthetic datasets, such as SAIL-VOS.

\section{Conclusion}

Perception is a crucial task for autonomous vehicles, yet current approaches still lack the amodal understanding necessary for the interpretation of complex traffic scenes.  To this regard, we presented AmodalSynthDrive, a multi-modal synthetic amodal perception dataset for autonomous driving. Using synthetic images and LiDAR point clouds, we provide a comprehensive dataset including ground truth annotations for fundamental amodal perception tasks while also introducing a new task for enhanced spatial understanding called amodal depth estimation. We provide over 60,000 individual image sets, each associated with amodal instance segmentation, amodal semantic segmentation, amodal panoptic segmentation, optical flow, 2D \& 3D bounding boxes, amodal depth as well as bird-eye-view maps. With AmodalSynthDrive, we provide various baselines and believe that this work will pave the way for novel research on amodal scene understanding of dynamic urban environments.


\footnotesize
\bibliographystyle{IEEEtran}
\bibliography{paper.bib}

\clearpage
\renewcommand{\baselinestretch}{1}
\setlength{\belowcaptionskip}{0pt}

\begin{strip}
\begin{center}
\vspace{-5ex}
\textbf{\LARGE \bf
AmodalSynthDrive: A Synthetic Amodal Perception Dataset for\\Autonomous Driving
}

\vspace{2ex}

\Large{\bf- Supplementary Material -}\\
\vspace{0.4cm}
\normalsize{Ahmed Rida Sekkat$^{*,1}$, Rohit Mohan$^{*,2}$,  Oliver Sawade$^{1}$, Elmar Matthes$^{1}$, and~Abhinav Valada$^2$}
\end{center}
\end{strip}

\setcounter{section}{0}
\setcounter{equation}{0}
\setcounter{figure}{0}
\setcounter{table}{0}
\setcounter{page}{1}
\makeatletter

\renewcommand{\thesection}{S.\arabic{section}}
\renewcommand{\thesubsection}{S.\arabic{section}.\arabic{subsection}}
\renewcommand{\thetable}{S.\arabic{table}}
\renewcommand{\thefigure}{S.\arabic{figure}}


In this supplementary material, we provide additional insights into various aspects of our work. We begin by discussing the architectural details of our proposed baselines for our proposed amodal depth estimation task in \secref{sec:baselines}. Then, we outline the training protocols for all the models benchmarked across the different amodal perception tasks within AmodalSynthDrive in \secref{sec:training}. Finally, in \secref{sec:doc}
we provide a comprehensive overview of our dataset, encompassing both the known issues and the datasheet.

\section{Amodal Depth Estimation Baselines}
\label{sec:baselines}

For the amodal depth estimation task, we establish two baselines derived by adapting the VIP-DeepLab~\cite{qiao2021vip} architecture. We develop these baselines such that they only retain the depth estimation components of the original network. The restructured network that we call ADB-Deeplab and Amodal-DORN, incorporates an encoder, a multi-scale context extractor, and a decoder equipped with a depth prediction head. The architecture of this baseline is depicted in \figref{fig:arch}. The encoder is based on the Resnet-50 architecture, while the multi-scale feature extractor is Dense Prediction Cell~\cite{chen2018searching} with 256 channels. The decoder utilizes skip connections and features two prediction heads. These heads correspond to modal depth estimation and amodal depth prediction. The amodal depth prediction head is set with a parameter $N$ equal to 8. 

Following the AD-Deeplab baseline, we introduce another baseline known as the ADS-Deeplab. In this baseline, we incorporate multi-headed attention feature propagation blocks across the various depth estimation heads. For each depth estimation head at level $l$, the modal depth head generates multiple key and value embeddings. Similarly, the fused features from up to the $l-1$ layer yield a distinct set of multiple key and value embeddings, tailored to that specific depth layer. The multi-headed attention blocks utilize the amodal features of the $l$th layer as their query. These features are then combined with the attended features from the multi-headed attention block, subsequently processed through a $5\times5$ depthwise separable convolution, and finally fed into the $1\times1$ depth prediction head. The architecture of AD-Deeplab is illustrated in~\figref{fig:arch}.

\section{Training Protocol}
\label{sec:training}

In this section, we describe the training protocol we employ for establishing baselines across various benchmarking tasks. We leverage PyTorch~\cite{paszke2019pytorch} for all of the baseline architectures. We train all the models on a system with an Intel Xenon (2.20GHz) processor and $8$ NVIDIA TITAN RTX GPUs with a batch size of $8$. We use a crop resolution of $1080\times1920$ pixels and use data augmentations such as random scaling which range from 0.5 to 2.0, as well as random flipping.

\subsection{Amodal Panoptic Segmentation}

{
\noindent\textbf{Amodal-EfficientPS~\cite{mohan2022amodal}}: We train AmodalEfficientPS using the stochastic gradient descent optimizer with a multi-step learning rate schedule where the drop factor is set to 10. We use a base learning rate of $0.01$ and train the model for 40 epochs. We set the milestones as $65\%$ and $90\%$ of the total epochs.
}

{\parskip=3pt
\noindent\textbf{APSNet~\cite{mohan2022amodal}}: The training process of APSNet spans 40 epochs where we use the stochastic gradient descent optimizer with an initial learning rate of $0.01$. Throughout the training, we employ a multi-step learning rate schedule, decreasing it by a factor of 10 at $65\%$ and $90\%$ of the total epochs.
}

{\parskip=3pt
\noindent\textbf{PAPS~\cite{mohan2022perceiving}}: We use a two-stage training protocol for PAPS. In each stage, we utilize the Adam optimizer along with a poly learning rate schedule, setting the initial learning rate at $0.001$. We train the model for a total of 300K iterations. We set N equal to 8 for the relative occlusion order layers.
Initially, we trained the model without the inclusion of the amodal mask refiner. After this, we keep the weights of the previous architectural components constant and focus solely on training the amodal mask refiner.
}

\subsection{Amodal Instance Segmentation}

\begin{figure*}
    \centering
        \includegraphics[width=\linewidth]{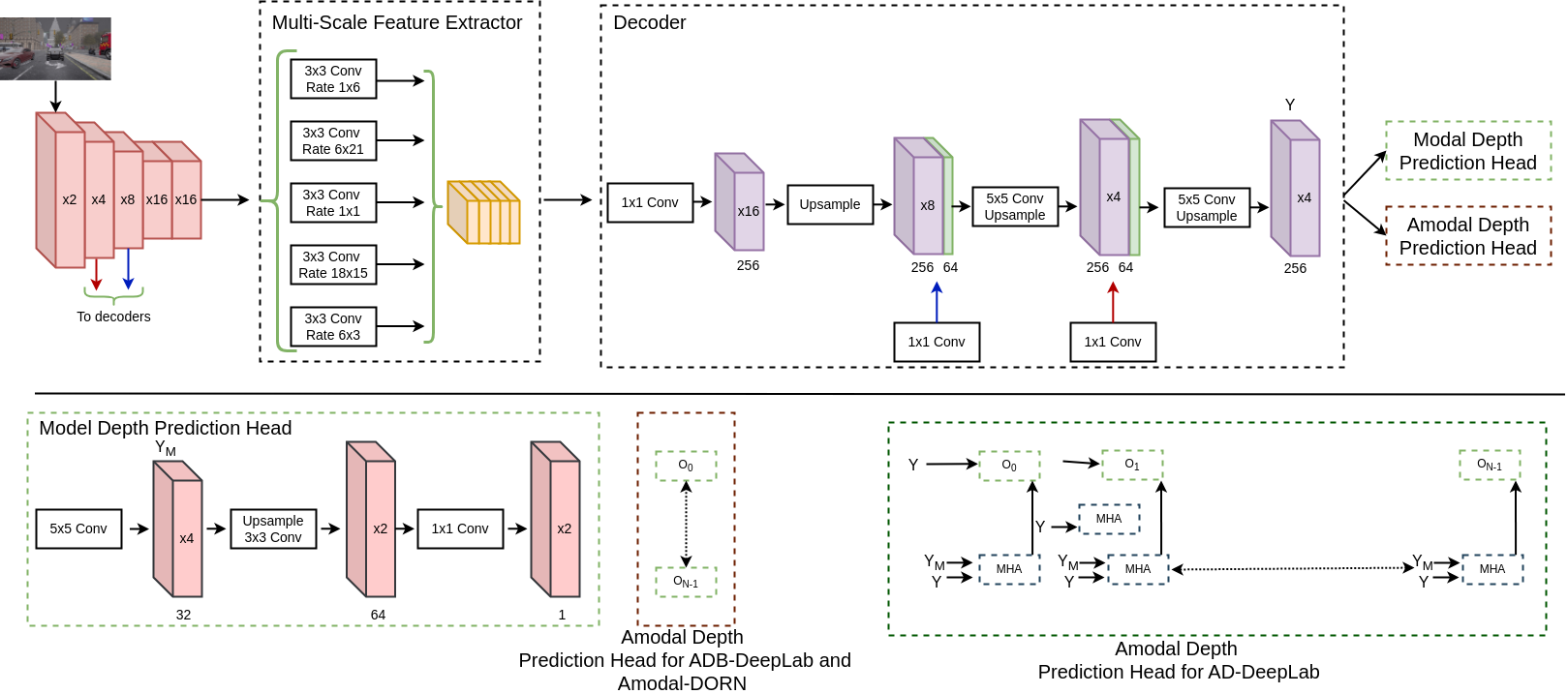}
    \caption{Architectural topology of the amodal depth estimation baseline: ADB-DeepLab, Amodal-DORN and AD-Deeplab. Please note that the boxes enclosed in colored dashes in each of the architecture correspond to the expanded version of the same colored boxes depicted on the bottom.}
    \label{fig:arch}
\end{figure*}

{
\noindent\textbf{ASN~\cite{qi2019amodal}}: To optimize the ASN model, we employ stochastic gradient descent with a learning rate of 0.01 and a momentum of 0.9. We train the model for 192K iterations. The learning rate is decayed subsequently by a factor of 0.1 at 130K and 180K iterations. 

}

{\parskip=3pt
\noindent\textbf{BCNet~\cite{ke2021deep}}: In order to train the BCNet model, we use stochastic gradient descent as the optimizer, with an initial learning rate of 0.01. We train this model for 192K iterations. We adjust the learning rate by decreasing it by a factor of 0.1 at 130K and 180K iterations.
}

{\parskip=3pt
\noindent\textbf{AISFormer~\cite{tran2022aisformer}}: For training the AISFormer model, we utilize stochastic gradient descent with an initial learning rate of 0.0025. We train for a total of 250K iterations. Further, we reduced the learning rate by a factor of 0.1 at two milestones (160K, 230K) during the training.
}

\subsection{Amodal Semantic Segmentation}

{
\noindent\textbf{amERFNet~\cite{breitenstein2022amodal}}: To train the amERFNet, we employ the Adam optimizer, beginning with a learning rate of 0.01. The model is trained over 40 epochs, during which we apply exponential decay to the learning rate.
}

{\parskip=3pt
\noindent\textbf{Y-ERFNet~\cite{breitenstein2023joint}}: To facilitate the training of Y-ERFNet, we use the Adam optimizer with an initial learning rate of 0.01. An exponential decay is applied to this learning rate as we train the model for over 40 epochs.
}

\subsection{Amodal Depth Estimation}
We resize the image to 540 × 960 resolution for training

{\parskip=3pt
\noindent\textbf{Amodal-DORN}: In order to train the Amodal-DORN model, we employ the Adam optimizer with an initial learning rate of 0.001. The learning rate is decayed exponentially throughout the training process, for a total of 120K iterations.
}

{\parskip=3pt
\noindent\textbf{ADB-DeepLab}: In order to train the AD-Deeplab model, we employ the Adam optimizer with an initial learning rate of 0.001. The learning rate is decayed exponentially throughout the training process, for a total of 120K iterations.
}

{\parskip=3pt
\noindent\textbf{AD-DeepLab}: For the training of the ADS-Deeplab model, we utilize the Adam optimizer, commencing with a learning rate of 0.001. We apply exponential decay to the learning rate during the training for a total of 120K iterations.
}

Please note the use of the terms "epochs" and "iterations" is due to the baseline models implemented in different frameworks.

\section{Dataset Documentation}
\label{sec:doc}

In this section, we highlight known issues and provide a datasheet for our AmodalSynthDrive dataset. Additionally, Table~\ref{tab:shapec} outlines the semantic classes present in the dataset and their corresponding class IDs. Moreover, Table~\ref{tab:weather} provides a comprehensive list of all the weather conditions encapsulated within the AmodalSynthDrive dataset.

\begin{table}
\centering\small\renewcommand\arraystretch{1}
\renewcommand{\tabcolsep}{2mm}

\begin{minipage}{0.48\textwidth}
\centering
\caption{Semantic Classes in AmodalSynthDrive}
\begin{tabular}{l|c|c|c}
\toprule
Name & ID & Attribute & ignoreInEval \\
\midrule
unlabeled  & 0   & - & True \\
road & 7   & stuff & False \\
sidewalk & 8   & stuff & False \\
building & 11  & stuff & False \\
wall & 12  & stuff & False \\
fence & 13  & stuff & False \\
pole & 18  & stuff & False \\
traffic light & 19  & stuff & False \\
traffic sign & 20  & stuff & False \\
vegetation & 21  & stuff & False \\
terrain & 22  & stuff & False \\
sky & 23  & stuff & False \\
person & 24  & thing & False \\
car & 26  & thing & False \\
truck & 27  & thing & False \\
bus & 28  & thing & False \\
motor & 32  & thing & False \\
bike & 33  & thing & False \\
road line  & 34   & - & True \\
other  & 35   & - & True \\
water  & 36  & - & True \\
\bottomrule
\end{tabular}
\label{tab:shapec}
\end{minipage}
\hfill
\begin{minipage}{0.5\textwidth}
\centering
\caption{Weather Conditions in AmodalSynthDrive}
\begin{tabular}{l}
\toprule
Weather Condition \\
\midrule
Clear Noon \\
Clear Sunset \\    
Cloudy Noon    \\  
Cloudy Sunset    \\
Default \\
Dust Storm    \\   
Hard Rain Noon \\
Hard Rain Sunset \\ 
Mid Rain Sunset  \\ 
Mid Rainy Noon   \\
Soft Rain Noon   \\
Soft Rain Sunset \\
Wet Cloudy Noon  \\
Wet Cloudy Sunset \\
Wet Noon \\
Wet Sunset \\
\bottomrule
\end{tabular}
\label{tab:weather}
\end{minipage}
\end{table}

\subsection{Known Issues}
\label{sec:known}

The AmodalSynthDrive dataset contains sequences of synchronized images from multiple cameras, as shown in the attached demo video where we highlight the ground truth (GT) annotations for the different tasks. It is worth noting that we do have two known issues related to the used simulator.

The first issue is related to the 3D bounding boxes of two-wheelers. In the version of the CARLA simulator that we used, the extent of the 3D bounding boxes of this type of vehicle is always equal to zero as discussed in issues number 3801 and 3670 in the official CARLA simulator repository on GitHub. At the time of writing this issue is not yet fixed, however, a post-generation fix will be performed by defining default extent values for each two-wheeler 3D model used.

The second problem pertains to the animations of both, the pedestrians in motion, and vehicles' wheels in movement. Currently, the pedestrians simply shift their positions and rotations without any animated transitions and the vehicles' wheels do not rotate when the vehicles are moving. Unfortunately, this is a limit to the current data generation method. If a resolution is discovered, we will make the necessary adjustments accordingly.

\myworries{\subsection{Datasheet for AmodalSynthDrive}}

\myworries{In this section, we present the datasheet for our dataset, adhering to the format recommended in~\cite{gebru2021datasheets}.}

\myworries{\subsubsection{Motivation}}

{
\noindent\textbf{\myworries{For what purpose was the dataset created?}}
\myworries{The dataset was created to facilitate research in amodal perception tasks, namely, amodal panoptic segmentation, amodal instance segmentation, amodal semantic segmentation, and amodal depth estimation. Please refer to the main paper for further details.}}

{\parskip=3pt
\noindent\textbf{\myworries{Who created the dataset?}}
\myworries{The authors are solely responsible for creating the dataset.}}

{\parskip=3pt
\noindent\textbf{\myworries{Who funded the creation of the dataset?}}
\myworries{Anonymized for the review}}

{\parskip=3pt
\noindent\textbf{\myworries{Any other comments?}}
\myworries{None.}}

\myworries{\subsubsection{Composition}}

{
\noindent\textbf{\myworries{What do the instances that comprise the dataset represent}}
\myworries{Urban driving scenarios featuring both static elements such as buildings, roads, and sky, as well as dynamic traffic participants including cars, pedestrians, etc.}}

{\parskip=3pt
\noindent\textbf{\myworries{How many instances are there in total?}}
\myworries{The dataset comprises a total of 2.8 million object annotations.}}

{\parskip=3pt
\noindent\textbf{\myworries{Does the dataset contain all possible instances or is it a sample (not
necessarily random) of instances from a larger set?}}
\myworries{The dataset contains all possible instances.}}

{\parskip=3pt
\noindent\textbf{\myworries{What data does each instance consist of?}}
\myworries{Each instance of the frame consists of a color image, LiDAR scan, depth map, BEV map, optical flow, annotations for both modal and amodal perception tasks.}}

{\parskip=3pt
\noindent\textbf{\myworries{Is there a label or target associated with each instance?}}
\myworries{Each instance of an object is associated with one of the semantic classes listed in ~\ref{tab:shapec}}}

{\parskip=3pt
\noindent\textbf{\myworries{Is any information missing from individual instances?}}
\myworries{Everything is included. No data is missing}}

{\parskip=3pt
\noindent\textbf{\myworries{Are relationships between individual instances made explicit?}}
\myworries{None.}}

{\parskip=3pt
\noindent\textbf{\myworries{Are there recommended data splits?}}
\myworries{Yes.}}

{\parskip=3pt
\noindent\textbf{\myworries{Are there any errors, sources of noise, or redundancies in the
dataset?}}
\myworries{Refer~\secref{sec:known}.}}

{\parskip=3pt
\noindent\textbf{\myworries{Is the dataset self-contained, or does it link to or otherwise rely on
external resources?}}
\myworries{The dataset is entirely self-contained.}}

{\parskip=3pt
\noindent\textbf{\myworries{Does the dataset contain data that might be considered confidential
(e.g., data that is protected by legal privilege or by doctor–patient
confidentiality, data that includes the content of individuals’ nonpublic communications)?}}
\myworries{No.}}

{\parskip=3pt
\noindent\textbf{\myworries{Does the dataset contain data that, if viewed directly, might be offensive, insulting, threatening, or might otherwise cause anxiety?}}
\myworries{No.}}

{\parskip=3pt
\noindent\textbf{\myworries{Does the dataset identify any subpopulations (e.g., by age, gender)?}}
\myworries{No.}}

{\parskip=3pt
\noindent\textbf{\myworries{Is it possible to identify individuals (i.e., one or more natural persons), either directly or indirectly (i.e., in combination with other
data) from the dataset?}}
\myworries{No.}}

{\parskip=3pt
\noindent\textbf{\myworries{Does the dataset contain data that might be considered sensitive in
any way (e.g., data that reveals race or ethnic origins, sexual orientations, religious beliefs, political opinions or union memberships, or
locations; financial or health data; biometric or genetic data; forms of
government identification, such as social security numbers; criminal
history)?}}
\myworries{No.}}

{\parskip=3pt
\noindent\textbf{\myworries{Any other comments?}}
\myworries{None.}}

\myworries{\subsubsection{Collection Process}}

{
\noindent\textbf{\myworries{How was the data associated with each instance acquired?}}
\myworries{Please refer to Section 3.1 of our main paper.}}

{\parskip=3pt
\noindent\textbf{\myworries{What mechanisms or procedures were used to collect the data (e.g.,
hardware apparatuses or sensors, manual human curation, software
programs, software APIs)?}}
\myworries{The CARLA simulator with the Python API was used to collect the data.}}

{\parskip=3pt
\noindent\textbf{\myworries{If the dataset is a sample from a larger set, what was the sampling strategy (e.g., deterministic, probabilistic with specific sampling probabilities)?}}
\myworries{N/A.}}

{\parskip=3pt
\noindent\textbf{\myworries{Who was involved in the data collection process (e.g., students,
crowdworkers, contractors) and how were they compensated (e.g.,
how much were crowdworkers paid)?}}
\myworries{Only the authors were involved in the data collection process.}}

{\parskip=3pt
\noindent\textbf{\myworries{Over what timeframe was the data collected?}}
\myworries{8 months.}}

{\parskip=3pt
\noindent\textbf{\myworries{Were any ethical review processes conducted (e.g., by an institutional review board)?}}
\myworries{N/A.}}

{\parskip=3pt
\noindent\textbf{\myworries{Did you collect the data from the individuals in question directly, or
obtain it via third parties or other sources (e.g., websites)?}}}
\myworries{N/A.}

{\parskip=3pt
\noindent\textbf{\myworries{Were the individuals in question notified about the data collection?}}
\myworries{N/A.}}

{\parskip=3pt
\noindent\textbf{\myworries{If consent was obtained, were the consenting individuals provided
with a mechanism to revoke their consent in the future or for certain
uses?}}
\myworries{N/A.}}

{\parskip=3pt
\noindent\textbf{\myworries{Has an analysis of the potential impact of the dataset and its use
on data subjects (e.g., a data protection impact analysis) been conducted?}}
\myworries{N/A.}}

{\parskip=3pt
\noindent\textbf{\myworries{Any other comments?}}
\myworries{None.}}

\myworries{\subsubsection{Preprocessing/cleaning/labeling}}
{\parskip=3pt
\noindent\textbf{\myworries{Was any preprocessing/cleaning/labeling of the data done (e.g., discretization or bucketing, tokenization, part-of-speech tagging, SIFT
feature extraction, removal of instances, processing of missing values)?}}
\myworries{No.}}

{\parskip=3pt
\noindent\textbf{\myworries{Was the “raw” data saved in addition to the preprocessed/cleaned/labeled data (e.g., to support unanticipated
future uses)?}}
\myworries{N/A.}}

{\parskip=3pt
\noindent\textbf{\myworries{Is the software that was used to preprocess/clean/label the data available? If so, please provide a link or other access point.}}
\myworries{No.}}

{\parskip=3pt
\noindent\textbf{\myworries{Any other comments?}}
\myworries{None.}}

\myworries{\subsubsection{Uses}}
{\parskip=3pt
\noindent\textbf{\myworries{Has the dataset been used for any tasks already?}}
\myworries{No.}}

{\parskip=3pt
\noindent\textbf{\myworries{Is there a repository that links to any or all papers or systems that
use the dataset?}}
\myworries{N/A.}}

{\parskip=3pt
\noindent\textbf{\myworries{What (other) tasks could the dataset be used for?}}
\myworries{Multi-modal variations of fundamental amodal perception tasks. Derivative information from our dataset can also be used to formulate novel tasks.}}

{\parskip=3pt
\noindent\textbf{\myworries{Is there anything about the composition of the dataset or the way it
was collected and preprocessed/cleaned/labeled that might impact
future uses?}}
\myworries{No.}}

{\parskip=3pt
\noindent\textbf{\myworries{Are there tasks for which the dataset should not be used?}}
\myworries{N/A.}}

{\parskip=3pt
\noindent\textbf{\myworries{Any other comments?}}
\myworries{None.}}

\myworries{\subsubsection{Distribution}}
{\parskip=3pt
\noindent\textbf{\myworries{Will the dataset be distributed to third parties outside of the entity (e.g., company, institution, organization) on behalf of which the
dataset was created?}}
\myworries{N/A.}}

{\parskip=3pt
\noindent\textbf{\myworries{How will the dataset will be distributed (e.g., tarball on website, API,
GitHub)?}}
\myworries{The dataset is available at \url{http://amodalsynthdrive.cs.uni-freiburg.de}}}

{\parskip=3pt
\noindent\textbf{\myworries{When will the dataset be distributed?}}
\myworries{The dataset is online since June, 2023}}

{\parskip=3pt
\noindent\textbf{\myworries{Will the dataset be distributed under a copyright or other intellectual
property (IP) license, and/or under applicable terms of use (ToU)?}}
\myworries{Yes. Our dataset is published under Non-Commercial Use Only license. Please refer to the accompanying license for details.}}

{\parskip=3pt
\noindent\textbf{\myworries{Have any third parties imposed IP-based or other restrictions on the
data associated with the instances?}}
\myworries{No.}}

{\parskip=3pt
\noindent\textbf{\myworries{Do any export controls or other regulatory restrictions apply to the
dataset or to individual instances?} }
\myworries{No.}}

{\parskip=3pt
\noindent\textbf{\myworries{Any other comments?}}
\myworries{None.}}

\myworries{\subsubsection{Maintenance}}
{\parskip=3pt
\noindent\textbf{\myworries{Who will be supporting/hosting/maintaining the dataset?}}
\myworries{The AmodalSynthDrive will be hosted permanently in the link \url{http://amodalsynthdrive.cs.uni-freiburg.de}. The maintenance will be ensured by the authors.}}

{\parskip=3pt
\noindent\textbf{\myworries{How can the owner/curator/manager of the dataset be contacted
(e.g., email address)?}}
\myworries{Anonymized for the review}}

{\parskip=3pt
\noindent\textbf{\myworries{Is there an erratum?}}
\myworries{No.}}

{\parskip=3pt
\noindent\textbf{\myworries{Will the dataset be updated (e.g., to correct labeling errors, add new
instances, delete instances)?}}
\myworries{Yes. Adding new instances with new environment releases of the CARLA simulator.}}

{\parskip=3pt
\noindent\textbf{\myworries{If the dataset relates to people, are there applicable limits on the retention of the data associated with the instances (e.g., were the individuals in question told that their data would be retained for a fixed
period of time and then deleted)?}}
\myworries{N/A.}}

{\parskip=3pt
\noindent\textbf{\myworries{Will older versions of the dataset continue to be supported/hosted/maintained?}}
\myworries{N/A.}}

{\parskip=3pt
\noindent\textbf{\myworries{If others want to extend/augment/build on/contribute to the dataset,
is there a mechanism for them to do so? }}
\myworries{Yes they can reach out to the authors.}}

{\parskip=3pt
\noindent\textbf{\myworries{Any other comments?}}
\myworries{None.}}

\end{document}